\documentclass[conference,onecolumn]{IEEEtran}
\IEEEoverridecommandlockouts
\usepackage{cite}
\usepackage{amsmath,amssymb,amsfonts}
\usepackage{graphicx}
\usepackage{textcomp}
\usepackage{xcolor}
\usepackage{tikz}
\usetikzlibrary{arrows.meta,positioning,calc,shapes.geometric,fit,backgrounds}
\usepackage{float}
\usepackage{placeins}
\usepackage{array}
\usepackage{booktabs}
\usepackage{multirow}
\usepackage{tabularx}
\usepackage{ragged2e}
\usepackage{url}
\newcolumntype{L}[1]{>{\RaggedRight\arraybackslash}p{#1}}
\newcolumntype{Y}{>{\RaggedRight\arraybackslash}X}

\def\BibTeX{{\rm B\kern-.05em{\sc i\kern-.025em b}\kern-.08em
    T\kern-.1667em\lower.7ex\hbox{E}\kern-.125emX}}
\definecolor{boxfill}{RGB}{226,238,250}
\definecolor{boxedge}{RGB}{42,91,145}
\definecolor{accentfill}{RGB}{255,235,204}
\definecolor{accentedge}{RGB}{204,112,0}

\tikzset{
procbox/.style={
rectangle,
rounded corners=2pt,
draw=boxedge,
thick,
fill=boxfill,
align=center,
inner sep=4pt,
minimum height=11mm,
minimum width=48mm,
text width=44mm,
font=\footnotesize
},
databox/.style={
rectangle,
draw=boxedge,
thick,
fill=boxfill,
align=center,
inner sep=4pt,
minimum height=9mm,
minimum width=30mm,
text width=27mm,
font=\footnotesize
},
accentbox/.style={
rectangle,
rounded corners=2pt,
draw=accentedge,
very thick,
fill=accentfill,
align=center,
inner sep=4pt,
minimum height=11mm,
minimum width=48mm,
text width=44mm,
font=\footnotesize
},
flow/.style={
-{Stealth[length=2.4mm,width=1.6mm]},
thick,
draw=boxedge
},
edgelbl/.style={
font=\scriptsize,
midway,
fill=white,
inner sep=1.5pt,
rounded corners=1pt
}
}

\begin{document}
\title{ToolMenuBench: Benchmarking Tool-Menu Filtering Strategies for Reliable and Efficient LLM Agents}

\author{\IEEEauthorblockN{Rahul Suresh Babu}
\IEEEauthorblockA{\textit{Independent Researcher} \\
United States of America \\
rahulsb@bu.edu}
\and
\IEEEauthorblockN{Laxmipriya Ganesh Iyer}
\IEEEauthorblockA{\textit{Independent Researcher} \\
United States of America \\
iyer.la@northeastern.edu}
}

\maketitle

\begin{abstract}
Tool-augmented large language model agents increasingly operate over large tool libraries, but existing evaluations often focus on whether a model can call a tool correctly rather than how the visible tool menu shapes reliability, efficiency, and safety-relevant risk exposure. We introduce ToolMenuBench, a benchmark for evaluating tool-menu construction in multi-step LLM agents. ToolMenuBench varies tool-menu size, distractor type, state-dependent task structure, and risk exposure, and reports both filter-level and downstream agent metrics, including visible-tool count, risky-tool exposure, task success, wrong-tool calls, premature actions, and token usage. In a controlled evaluation across seven model backends, three tool-menu sizes, six filtering methods, and seven evaluation settings, CMTF improves task success from 32.1\% under all-tools exposure to 85.7\%, while reducing average token usage by roughly 98\%. Causal minimal tool filtering achieves the strongest overall tradeoff, reducing visible tools, wrong-tool calls, premature actions, and risky-tool exposure relative to unfiltered exposure, lexical filtering, state-aware filtering, and broader causal-path baselines. ToolMenuBench provides a reusable evaluation framework for studying the agent-interface problem: which tools should be visible, when they should be visible, and under what cost or risk constraints.
\end{abstract}

\begin{IEEEkeywords}
Tool-augmented LLM agents, tool-menu construction, tool filtering, agent benchmarking, function calling, causal tool filtering, preconditions and effects, tool retrieval, LLM reliability, agent safety
\end{IEEEkeywords}

\section{Introduction}
\label{sec:introduction}

Tool-augmented large language model (LLM) agents increasingly rely on external tools to search information, access files, update calendars, draft emails, execute code, query databases, and interact with structured services~\cite{yao2022react,schick2023toolformer,qin2023toollm,li2023apibank}. As these agents move from small demonstrations to larger tool ecosystems, the visible tool menu becomes a central part of the agent interface. A model may be capable of calling an individual tool correctly, yet still fail when it must choose among many semantically similar, partially overlapping, premature, risky, or irrelevant tools.

Existing tool-use benchmarks have made substantial progress in evaluating function calling, API use, argument generation, and multi-step tool execution~\cite{qin2023toollm,li2023apibank,patil2025bfcl}. However, many evaluations treat the available tool set as fixed, curated, or given. They do not fully isolate the tool-menu design problem: which tools should be visible to the agent at each decision step, how large the visible menu should be, how distractors affect reliability, and how filtering should trade off task success, token cost, and safety-relevant risk exposure.

This distinction matters because tool exposure is not merely a prompt-length issue. Exposing all tools gives the agent maximum flexibility, but it also increases the chance of wrong-tool calls, premature actions, and unnecessary token usage. Semantic retrieval and tool-pruning methods can reduce menu size, but semantically relevant tools are not always causally appropriate for the current task state~\cite{shi2025toolret,liu2025toolscope}. For example, a calendar task may contain tools for searching events, creating events, updating events, deleting events, and sending notifications. Many of these tools are domain-relevant, but only one may be the correct next action given the current state and goal.

Recent work on causal tool filtering argues that tool menus should be shaped by preconditions, effects, and task-state progress rather than semantic relevance alone~\cite{babu2026toolchoice}. Follow-up work on contract learning studies how such precondition--effect contracts can be inferred from schemas, documentation, and execution traces~\cite{babu2026contract2tool}. These methods suggest that reliable agents require more than better tool-call syntax: they require systematic evaluation of the interface that decides what the model is allowed to see and choose.

We introduce \textit{ToolMenuBench}, a benchmark for evaluating tool-menu construction in multi-step LLM agents. ToolMenuBench makes the visible tool menu itself the object of evaluation. It tests how agents behave as tool libraries grow, as distractors become more realistic, and as filtering methods impose different notions of relevance, executability, causal necessity, efficiency, and risk exposure. The benchmark supports both filter-level evaluation, which measures what the agent is allowed to see before acting, and downstream agent evaluation, which measures task success, wrong-tool calls, premature actions, token usage, and risky-tool exposure during execution.

In this paper, we report a core controlled evaluation of six implemented filtering methods: all-tools exposure, keyword top-$5$ filtering, keyword top-$10$ filtering, state-aware filtering, full causal path exposure, and causal minimal tool filtering (CMTF). ToolMenuBench is also designed to support additional extensions, including embedding-based retrieval, schema-aware filtering, learned-contract filtering, recovery-menu filtering, and cost- or risk-aware filtering. The broader goal is to make tool-menu design a first-class evaluation target for tool-augmented agents.

\noindent\textbf{Contributions.} This paper makes the following contributions:
\begin{enumerate}
\item We introduce ToolMenuBench, a benchmark for evaluating how visible tool-menu construction affects reliability, efficiency, and safety-relevant exposure in multi-step LLM agents.
\item We define a distractor taxonomy covering semantic distractors, near-duplicate tools, schema-compatible wrong tools, premature tools, risky tools, and cross-domain distractors.
\item We define filter-level and downstream metrics for tool-menu evaluation, including visible-tool count, risky-tool exposure, task success, wrong-tool calls, premature actions, and token usage.
\item We report a controlled core evaluation across multiple model backends, tool-menu sizes, filtering methods, and distractor settings, showing that causally aligned menus can improve success while reducing tool exposure and token usage.
\item We provide benchmark artifacts including task definitions, tool registries, gold trajectories, tool contracts, distractor annotations, filtering baselines, evaluation scripts, and result-generation utilities.
\end{enumerate}

\section{Background and Related Work}
\label{sec:background}

\subsection{Tool-Augmented LLM Agents}

Tool use has become a central mechanism for extending large language models beyond text generation. ReAct introduced interleaved reasoning and acting, allowing models to combine language-based reasoning with external actions~\cite{yao2022react}. Toolformer showed that language models can learn to call external APIs during inference~\cite{schick2023toolformer}. ToolLLM and ToolBench scaled this setting to large collections of real-world APIs and multi-step tool-use tasks~\cite{qin2023toollm}, while API-Bank evaluated tool-augmented dialogue and API use in controlled settings~\cite{li2023apibank}. These works establish tool use as a core agent capability, often under a fixed, curated, or task-provided tool set. ToolMenuBench studies a complementary systems question: how should the visible tool menu be constructed when many tools are available?

\subsection{Function Calling and Tool-Use Benchmarks}

Function-calling benchmarks evaluate whether models can produce valid and useful tool calls. API-Bank, ToolBench, and the Berkeley Function Calling Leaderboard measure capabilities such as tool selection, argument construction, multi-turn function calling, and agentic tool use~\cite{li2023apibank,qin2023toollm,patil2025bfcl}. These benchmarks are essential for measuring model-level tool-use capability, but they often assume that the candidate tool set has already been selected. A model tested with a small curated tool set may behave differently when exposed to a large registry containing near-duplicates, schema-compatible wrong tools, premature actions, and risky operations. ToolMenuBench isolates this interface variable by evaluating how different menu-construction strategies affect both filter-level and downstream agent behavior.

\subsection{Tool Retrieval and Semantic Filtering}

As tool libraries grow, retrieval and pruning methods are needed to reduce prompt size and ambiguity. Tool retrieval methods select candidate tools using lexical matching, embedding similarity, or learned retrieval models. Recent work shows that retrieval models are not always tool-savvy, even when they perform well on conventional retrieval tasks~\cite{shi2025toolret}. ToolScope addresses related scalability and ambiguity issues by merging overlapping tools and applying context-aware filtering~\cite{liu2025toolscope}. These approaches are useful, but they primarily treat filtering as a relevance or redundancy-reduction problem. ToolMenuBench is designed to evaluate retrieval-style filtering alongside state-aware and causal filtering because a tool can be semantically relevant while still being premature, unnecessary, or unsafe at the current step.

\subsection{State-Aware and Causal Tool Filtering}

State-aware filtering exposes tools whose required arguments or preconditions are satisfied by the current task state. This can reduce invalid calls, but executability is not the same as usefulness: a callable tool may not advance the current goal. Causal filtering represents tools with lightweight preconditions and effects, then exposes tools according to their role in moving the current state toward the goal~\cite{babu2026toolchoice}. Causal minimal tool filtering (CMTF) exposes only the next causally necessary frontier, while Contract2Tool studies how such precondition--effect contracts can be learned from schemas, documentation, and execution traces~\cite{babu2026contract2tool}. ToolMenuBench builds on these ideas by evaluating state-aware and causal filtering under a shared benchmark, while also defining benchmark extensions for learned-contract settings.

\subsection{Risk, Cost, and Realistic Distractors}

Realistic tool menus contain more than irrelevant tools. They often include near-duplicate APIs, schema-compatible wrong tools, premature actions, cross-domain distractors, and high-risk operations such as deleting files, sending messages, sharing documents, changing permissions, or modifying external state. Complementary work on self-healing agentic orchestrators studies runtime reliability after tool-use failures occur, including failure classification, budgeted recovery, and verifier-guided trajectory repair~\cite{babu2026selfhealing}. ToolMenuBench focuses on the earlier interface layer: which actions are visible before the model chooses. It explicitly models realistic distractors through a taxonomy and measures not only task success and token usage, but also gold next-tool exposure, extra tools exposed, risky-tool exposure, and unauthorized risky-tool exposure.

\subsection{Benchmark Gap}

Existing benchmarks have advanced evaluation of model-level function calling and multi-step API use, while retrieval, causal filtering, and runtime-reliability work have improved tool selection and recovery from different perspectives. What remains missing is a benchmark that treats the visible tool menu itself as the object of evaluation. ToolMenuBench addresses this gap by evaluating tool-menu construction across menu size, distractor type, task-state dependency, token cost, and risk exposure, while also supporting extensions for contract-quality and policy-aware evaluation. It frames tool-menu design as a first-class systems problem for reliable tool-augmented LLM agents.

\section{Benchmark Design}
\label{sec:benchmark_design}

ToolMenuBench is designed to evaluate tool-menu construction in multi-step LLM agents. The benchmark separates three questions that are often conflated in tool-use evaluation: whether the correct next tool is visible, whether the model selects it, and whether the resulting trajectory completes the task. To support this separation, ToolMenuBench includes a controlled tool registry, multi-step tasks, gold trajectories, precondition--effect contracts, distractor annotations, risk labels, optional cost labels, and evaluation scripts.

\subsection{Benchmark Overview}

Each benchmark instance consists of a user task, an initial symbolic state, a goal state, a tool registry, and a filtering method. At each decision step, the filtering method maps the task, current state, goal, and registry to a visible subset of tools. The agent receives the task, current state, and visible tool menu, then selects one tool call. A deterministic environment executes the selected tool, updates the symbolic state according to the tool effects, and returns an observation. The process continues until the goal state is reached or the step budget is exhausted.

This design supports both filter-level and downstream evaluation. Filter-level evaluation asks whether the filtering method exposes the gold next tool before the model acts. Downstream evaluation asks whether the model completes the task when operating under that filtered menu.

\subsection{Task Domains}

ToolMenuBench focuses on workflow domains where multi-step tool use is natural and where tool confusion can produce meaningful failures. The benchmark includes:
\begin{itemize}
\item \textit{Calendar workflows}: searching events, reading event details, creating events, updating events, deleting events, and sending event notifications.
\item \textit{Email workflows}: searching messages, reading messages, drafting replies, sending emails, forwarding messages, and applying labels.
\item \textit{File and document workflows}: locating files, reading documents, summarizing content, editing files, sharing documents, and deleting files.
\item \textit{Contact and identity workflows}: looking up contacts, resolving ambiguous recipients, and checking confirmation or authorization state.
\end{itemize}
These domains were chosen because they require ordered state transitions. For example, an agent may need to search for an event before updating it, read a file before summarizing it, or draft an email before sending it.

\subsection{Tool Registry}

The benchmark uses a synthetic but structured tool registry designed to isolate controlled tool-menu conditions. Each tool is represented by a name, natural-language description, input schema, output schema, required state variables, produced state variables, risk label, optional cost label, domain label, and distractor category when applicable.

The registry contains both task-relevant tools and distractor tools. Task-relevant tools are needed for at least one gold trajectory. Distractor tools are designed to be plausible but incorrect under some task states. This makes the benchmark sensitive to realistic tool-menu failures rather than only random irrelevant noise.

\subsection{Gold Trajectories and State Variables}

Each task includes a gold trajectory specifying the intended sequence of tools and the state variables produced after each step. Let $s_t$ denote the symbolic state before step $t$, and let $g$ denote the goal condition. A gold trajectory consists of a sequence:
\begin{equation}
(a_1, a_2, \ldots, a_k),
\end{equation}
where each action advances the state toward the goal by producing one or more task-relevant state variables.

These trajectories support two forms of evaluation. First, they define the gold next tool at each step for filter-level evaluation. Second, they define the expected state transitions used to determine whether the downstream agent successfully completes the task.

\subsection{Distractor Taxonomy}

ToolMenuBench defines a distractor taxonomy to evaluate why filtering methods fail. The main distractor categories are:
\begin{itemize}
\item \textit{Semantic distractors}: tools that sound relevant to the task but do not advance the current goal.
\item \textit{Near-duplicate tools}: tools with similar names or descriptions but different preconditions or effects.
\item \textit{Schema-compatible wrong tools}: tools that accept similar arguments but produce the wrong state transition.
\item \textit{Premature tools}: tools that may be useful later but should not be exposed at the current step.
\item \textit{Risky tools}: tools that perform high-impact actions such as sending, deleting, sharing, paying, or modifying external state.
\item \textit{Cross-domain distractors}: tools from unrelated domains that test broad menu noise.
\end{itemize}
This taxonomy makes it possible to evaluate not only whether a filtering strategy fails, but what kind of tool-menu ambiguity causes the failure.

\subsection{Benchmark Tracks}

ToolMenuBench is designed to support multiple benchmark tracks. The reported experiments in this paper focus on filtering strategy comparison, menu-size scaling, distractor robustness, efficiency, and risk exposure. Contract-quality sensitivity and full policy-aware cost/risk optimization are benchmark-supported extensions.

\subsubsection{Menu-Size Scaling}
This track evaluates how filtering methods behave as the available tool registry grows. The same tasks are evaluated under progressively larger tool menus by adding distractors while keeping the gold trajectory fixed.

\subsubsection{Distractor Robustness}
This track varies the type of distractors while holding the task structure fixed. It measures whether methods are vulnerable to semantic distractors, near-duplicates, schema-compatible wrong tools, premature tools, risky tools, or cross-domain tools.

\subsubsection{Filtering Strategy Comparison}
This track compares tool-menu construction methods under identical tasks and registries. The reported experiments evaluate all-tools exposure, keyword top-$5$, keyword top-$10$, state-aware filtering, full causal path exposure, and CMTF. The benchmark design also supports embedding-based retrieval, schema-aware filtering, and learned-contract CMTF.

\subsubsection{Contract-Quality Sensitivity}
This extension evaluates how filtering performance changes when contracts are gold, learned, incomplete, or corrupted. Contract perturbations include missing preconditions, spurious preconditions, missing effects, spurious effects, incorrect risk labels, and non-canonical state variables.

\subsubsection{Cost and Risk Awareness}
This extension evaluates filtering under token, tool-cost, and risk constraints. It measures whether a method can reduce token usage and risky-tool exposure while preserving task success, and whether high-risk tools can be hidden until required authorization or confirmation state is present.

\subsection{Benchmark Outputs}

For each run, ToolMenuBench records the visible tools at each step, risk labels for exposed tools, the model-selected tool, tool arguments, execution status, state updates, token usage, and final task outcome. These outputs support both per-step filter analysis and full-trajectory agent analysis. They allow researchers to distinguish failures caused by missing the gold tool, exposing too many plausible tools, selecting a wrong visible tool, taking a premature action, exposing risky tools, or exhausting the step budget.

\section{Filtering Methods}
\label{sec:filtering_methods}

ToolMenuBench compares tool-menu construction strategies that differ in how they decide which tools are visible at each decision step. Each method receives the same task, current symbolic state, goal condition, and tool registry, but exposes a different subset of tools to the downstream LLM agent. The reported experiments evaluate six core methods: all-tools exposure, keyword top-$5$, keyword top-$10$, state-aware filtering, full causal path exposure, and causal minimal tool filtering (CMTF). ToolMenuBench also supports additional filtering extensions, described at the end of this section.

\subsection{All-Tools Exposure}

The all-tools baseline exposes the full registry at every step. This guarantees that the gold tool is visible, but maximizes prompt length, token cost, risky-tool exposure, and tool-choice ambiguity. It serves as the reference point for measuring how much filtering reduces tool exposure.

\subsection{Keyword Top-$k$ Filtering}

Keyword top-$k$ filtering ranks tools by lexical overlap between the task and tool metadata, including the tool name, description, and optionally schema fields. The top-$k$ tools are exposed. This baseline is simple and inexpensive, but it is sensitive to wording and does not account for task state. The reported experiments evaluate $k=5$ and $k=10$.

\subsection{State-Aware Filtering}

State-aware filtering exposes tools whose preconditions are satisfied by the current symbolic state. Let $s_t$ denote the state at step $t$, and let $R_i$ denote the required state variables for tool $a_i$. A tool is visible if:
\begin{equation}
R_i \subseteq s_t.
\end{equation}
This is stronger than keyword filtering because it accounts for executability. However, executability is not equivalent to usefulness: a callable tool may still be irrelevant, risky, premature, or not on the next path to the goal.

\subsection{Full Causal Path Exposure}

Full causal path exposure uses preconditions and effects to expose tools that may appear on some valid path from the current state to the goal. This tests whether causal relevance is sufficient without next-step minimality. It can remove irrelevant tools, but may still expose future tools before they are needed.

\subsection{Causal Minimal Tool Filtering}

Causal Minimal Tool Filtering (CMTF) exposes only the next causally necessary frontier of tools. A tool is visible only when it is executable under the current state and its effects advance the remaining state toward the goal at the current step. This tests whether reliable tool menus should be shaped by causal necessity rather than semantic relevance, lexical overlap, or executability alone.

\subsection{Benchmark-Supported Extensions}

ToolMenuBench is designed to support additional filtering strategies beyond the core methods reported in this paper. Embedding top-$k$ filtering ranks tools by semantic similarity between the task and tool metadata. Schema-aware filtering exposes tools whose required input fields can be filled from the user request or current state. Learned-contract CMTF applies causal filtering using inferred precondition--effect contracts rather than manually specified contracts. Risk-aware filtering augments a base filter with policies that hide high-impact tools until they are causally relevant and required authorization or confirmation state is present. Cost-aware filtering incorporates token, latency, or execution-cost labels into menu construction. Recovery-menu filtering exposes recovery actions such as retry, verify, ask the user, use a fallback tool, escalate, or abort after tool failures. These extensions define the broader benchmark space and can be evaluated under the same ToolMenuBench protocol.

\subsection{Comparison Protocol}

All reported methods are evaluated under the same tasks, model backends, execution environment, prompting protocol, and metrics. This isolates the effect of the visible tool menu from model selection, task variation, and execution nondeterminism. By comparing unfiltered exposure, lexical filtering, executability-based filtering, broader causal-path filtering, and next-step causal minimal filtering, ToolMenuBench separates failures caused by tool-menu scale, missing or misleading visible tools, model behavior, and causal misalignment.

\section{Evaluation Metrics}
\label{sec:evaluation_metrics}

ToolMenuBench defines metrics at multiple levels so that failures can be attributed to the menu filter, the model's tool choice, or the resulting task trajectory. The reported experiments focus on filter-level and downstream agent metrics. Contract-level and explicit cost metrics are included as benchmark-supported extensions for learned-contract and policy-aware settings.

\subsection{Filter-Level Metrics}

Filter-level metrics are computed before the model selects a tool. They measure whether the visible menu contains the right action and how much unnecessary or risky exposure remains:
\begin{itemize}
\item \textit{Gold next-tool exposure}: whether the gold next tool is visible at the current step.
\item \textit{No-visible-tool rate}: how often the filter exposes no valid next tool.
\item \textit{Extra tools exposed}: the number of visible tools beyond the gold next tool.
\item \textit{Average visible tools}: the average tool-menu size per decision step.
\item \textit{Risky-tool exposure}: how often high-impact tools are visible.
\item \textit{Unauthorized risky-tool exposure}: how often high-impact tools are visible before required authorization or confirmation state is present.
\end{itemize}

\subsection{Downstream Agent Metrics}

Downstream metrics are computed over full task executions. They measure how the filtered menu affects actual agent behavior:
\begin{itemize}
\item \textit{Task success}: whether the task reaches the goal state within the step budget.
\item \textit{Wrong-tool calls}: the number of selected tools whose effects do not advance the current state toward the goal.
\item \textit{Premature actions}: the number of tools selected before their causal or authorization preconditions are satisfied.
\item \textit{Steps to completion}: the number of tool-use steps taken before success or termination.
\item \textit{Token usage}: the total prompt and completion tokens used during the task.
\item \textit{Cost per successful task}: an optional metric computed when token-cost or tool-cost labels are available.
\end{itemize}

\subsection{Contract-Level Extension Metrics}

For learned-contract or perturbed-contract tracks, ToolMenuBench can also evaluate intrinsic contract quality. Preconditions and effects can be scored using precision, recall, and F1 against gold contracts. Risk labels can be evaluated using classification accuracy. Learned-contract settings can additionally report exact contract match and invalid-output rate. These metrics help distinguish failures caused by contract quality from failures caused by filtering logic or model tool-choice behavior.

\subsection{Aggregate Reporting}

Results are aggregated by filtering method, model backend, tool-menu size, and distractor setting. This allows ToolMenuBench to identify not only which method performs best overall, but also which methods degrade under larger menus, near-duplicate tools, premature actions, risky tools, or specific model backends. The benchmark also supports domain-level and contract-quality analyses for future extensions.

\section{Experimental Setup}
\label{sec:experimental_setup}

We evaluate tool-menu construction under a controlled agent protocol in which the task, model backend, execution environment, and prompting format are held fixed while the visible tool menu is varied. The experimental matrix contains 26{,}460 end-to-end agent executions: seven evaluation settings, seven model backends, three tool-menu sizes, six filtering methods, and thirty tasks for each condition. This design isolates the effect of tool-menu construction on downstream task success, wrong-tool calls, premature actions, token usage, and risky-tool exposure.

\subsection{Models}

We evaluate seven model backends spanning multiple model families and capability levels: Nova 2 Lite, Nova Premier, Nova 2 Pro, Claude Haiku 4.5, Claude Sonnet 4.6, Claude Opus 4.8, and GPT-OSS-120B. Each model is evaluated under the same tasks, tool registries, menu sizes, and filtering methods. This allows us to report both aggregate trends across backends and model-level variation.

\subsection{Tasks, Menus, and Distractor Settings}

Each task is defined by a user request, an initial symbolic state, a goal state, and a gold tool trajectory. We evaluate thirty tasks for each combination of evaluation setting, model backend, menu size, and filtering method. To test menu-size scaling, we vary the available tool registry across three sizes: 25, 100, and 250 tools. The gold trajectory is held fixed while distractor tools are added to increase menu size.

We evaluate seven settings in total: one main mixed-distractor benchmark and six targeted distractor stress tests. The targeted settings isolate semantic distractors, near-duplicate tools, schema-compatible wrong tools, premature tools, risky tools, and cross-domain distractors. This design allows us to measure both overall filtering performance and robustness to specific classes of tool-menu ambiguity.

\subsection{Prompting Protocol}

At each decision step, the model receives the user task, the current symbolic state, and the schemas and descriptions for the tools exposed by the filtering method. The model is instructed to produce exactly one tool call with valid arguments or stop when the task is complete. The prompt format is held fixed across filtering methods; only the visible tool menu changes.

\subsection{Execution Environment}

We use a deterministic mocked tool environment. Each selected tool either updates the symbolic state according to its defined effects or produces a failure observation when its requirements are not satisfied. A run terminates when the goal state is reached or the step budget is exhausted. This controlled execution environment isolates tool-selection behavior from external API nondeterminism, network failures, authentication errors, rate limits, and other deployment-specific effects.

\subsection{Compared Methods}

In the reported experiments, we compare six tool-menu construction strategies: all-tools exposure, keyword top-$5$ filtering, keyword top-$10$ filtering, state-aware filtering, full causal path exposure, and CMTF. These methods span unfiltered exposure, lexical retrieval, executability-based filtering, broader causal-path filtering, and next-step causal minimal filtering.

ToolMenuBench is designed to support additional extensions, including embedding-based retrieval, schema-aware filtering, learned-contract CMTF, risk-aware filtering, cost-aware filtering, and recovery-menu filtering. We treat these as benchmark-supported extensions and leave their full empirical evaluation to future work.

\subsection{Reporting Protocol}

For each run, we record the visible tools, selected tool, arguments, execution result, state update, token usage, risky-tool exposure, and final task outcome. We store both per-task metrics and raw execution traces. Results are aggregated by filtering method, model backend, tool-menu size, and distractor setting. This reporting protocol allows us to distinguish failures caused by broad tool exposure, missing or misleading tool menus, model selection errors, premature actions, and risky-tool exposure.

\FloatBarrier

\section{Results}
\label{sec:results}

We evaluate ToolMenuBench along three axes: filtering strategy, tool-menu size, and distractor type. The main benchmark uses the mixed-distractor setting and evaluates seven model backends across three menu sizes and six filtering methods. We additionally evaluate six targeted distractor stress tests: semantic, near-duplicate, schema-compatible, premature, risky, and cross-domain distractors. Unless otherwise stated, success is reported in percent, average visible tools are measured per decision step, and the remaining metrics are averaged per task execution.

\subsection{Overall Filtering Strategy Comparison}
\label{sec:results-overall}

Table~\ref{tab:main-results} summarizes the main mixed-distractor benchmark. The results show that tool-menu construction has a large effect on downstream agent behavior. Exposing all tools achieves only 32.1\% success while requiring an average of 125.00 visible tools per step and 56{,}062 tokens per task execution. Keyword filtering substantially reduces token usage, but success remains low, indicating that lexical relevance alone is not sufficient for reliable tool use. State-aware and full-causal-path filtering eliminate premature actions in this setting, but they still expose many tools and do not improve success over broader baselines.

\begin{figure}[!t]
\centering
\includegraphics[width=0.78\linewidth]{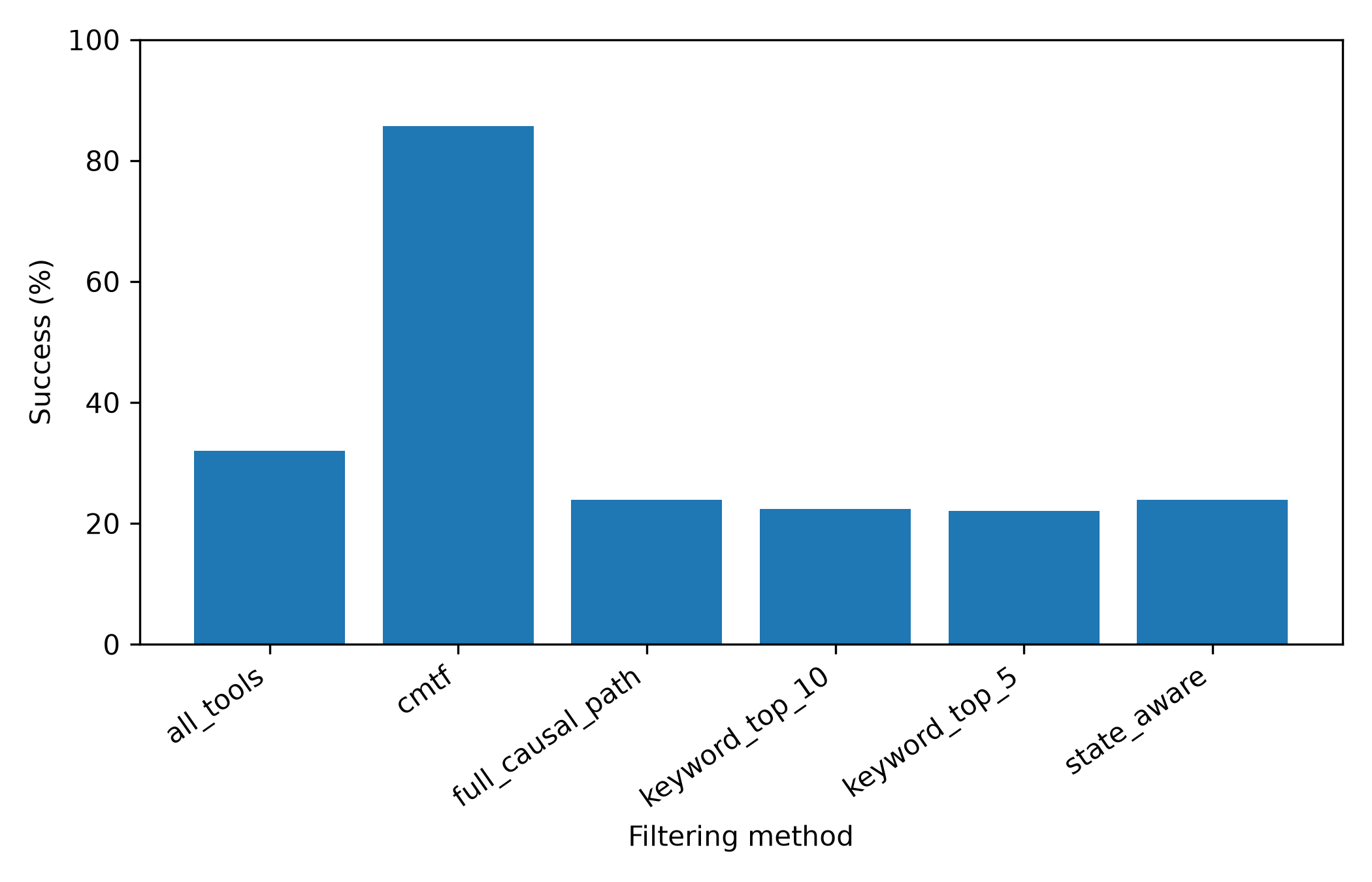}
\caption{Task success by filtering method in the main mixed-distractor benchmark. CMTF achieves the highest success among the evaluated filtering strategies.}
\label{fig:success-by-method}
\end{figure}

CMTF provides the strongest overall tradeoff. It achieves 85.7\% success while reducing the average visible menu to 0.99 tools and average token usage to 1{,}125 tokens. Compared with all-tools exposure, this is a 53.6 percentage-point absolute improvement in success and roughly a 98\% reduction in token usage. Figure~\ref{fig:success-by-method} visualizes the same comparison. These results support the central claim of ToolMenuBench: the visible tool menu changes the agent's decision problem, and reliable tool use requires more than exposing all possible actions.

\begin{table}[t]
\centering
\caption{Main mixed-distractor results by filtering method. Success is reported in percent. Average visible tools are measured per decision step; other metrics are averaged per task execution.}
\label{tab:main-results}
\begin{tabular}{lrrrrrr}
\toprule
Method & Success (\%) & Wrong tools & Premature & Avg. visible & Tokens & Risky exposed \tabularnewline
\midrule
All tools & 32.1 & 2.78 & 0.04 & 125.00 & 56{,}062 & 206.87 \tabularnewline
Keyword top-5 & 22.1 & 2.56 & 0.87 & 4.80 & 3{,}200 & 7.22 \tabularnewline
Keyword top-10 & 22.4 & 2.59 & 0.63 & 9.54 & 5{,}356 & 13.43 \tabularnewline
State-aware & 24.0 & 3.40 & 0.00 & 25.95 & 13{,}522 & 41.31 \tabularnewline
Full causal path & 24.0 & 3.40 & 0.00 & 26.24 & 13{,}697 & 41.53 \tabularnewline
CMTF & 85.7 & 0.53 & 0.00 & 0.99 & 1{,}125 & 0.71 \tabularnewline
\bottomrule
\end{tabular}
\end{table}

\subsection{Scaling with Tool-Menu Size}
\label{sec:results-menu-size}

\begin{figure}[!t]
\centering
\includegraphics[width=0.78\linewidth]{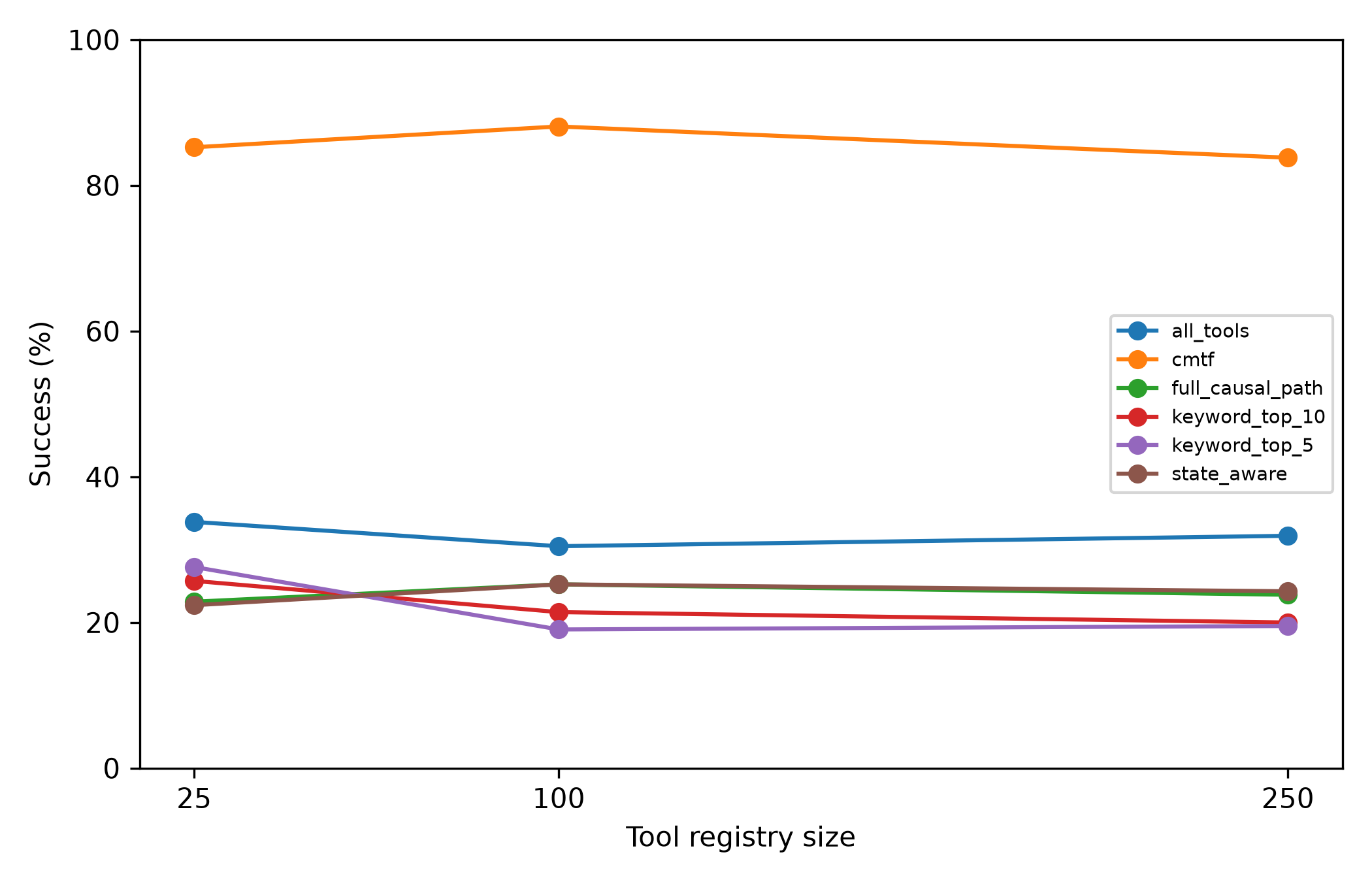}
\caption{Task success as the tool registry grows from 25 to 100 and 250 tools. Causal filtering remains comparatively stable as the available tool registry grows.}
\label{fig:success-vs-menu-size}
\end{figure}

The main benchmark varies the tool registry size from 25 to 100 and 250 tools while holding the task set and gold trajectories fixed. This isolates the menu-scaling problem: as the tool ecosystem grows, methods that expose broad menus impose higher token cost and greater tool-choice ambiguity. All-tools exposure is especially inefficient because every additional tool is included in the model context regardless of whether it is useful for the current state. In contrast, CMTF is largely insensitive to raw registry size because visibility is determined by the next causally necessary action rather than by the number of available tools. Figure~\ref{fig:success-vs-menu-size} shows that CMTF remains stable as the tool registry grows, while all-tools exposure and keyword-based filters remain substantially lower.

\subsection{Distractor Robustness}
\label{sec:results-distractors}

Table~\ref{tab:distractor-results} reports robustness on six targeted distractor stress tests. We exclude the mixed setting from this table because the mixed-distractor benchmark is already reported in Table~\ref{tab:main-results}. The targeted distractor settings test different failure modes: semantic distractors test surface-level relevance, near-duplicate tools test fine-grained tool ambiguity, schema-compatible distractors test argument-compatible wrong actions, premature distractors test exposure of future actions, risky distractors test high-impact actions, and cross-domain distractors test broad irrelevant noise.

For compactness, Table~\ref{tab:distractor-results} reports the best-performing method in each setting; all targeted distractor runs were evaluated across the same six filtering methods. Across all six targeted distractor types, CMTF is selected as the best method by success rate, with token usage used as a tie-breaker. Its success ranges from 84.4\% on premature distractors to 90.6\% on near-duplicate distractors, while maintaining low wrong-tool counts and approximately one thousand tokens per task execution. This pattern suggests that causal next-step filtering is robust not only to irrelevant tools, but also to realistic distractors that are semantically plausible, schema-compatible, premature, or risky.

\begin{table}[t]
\centering
\caption{Distractor robustness results. For each distractor type, the table reports the best method selected by highest success rate, with lower token usage used as a tie-breaker.}
\label{tab:distractor-results}
\begin{tabular}{llrrr}
\toprule
Distractor type & Best method & Success (\%) & Wrong tools & Tokens \tabularnewline
\midrule
Semantic & CMTF & 89.2 & 0.44 & 1{,}108 \tabularnewline
Near-duplicate & CMTF & 90.6 & 0.42 & 1{,}111 \tabularnewline
Schema-compatible & CMTF & 89.8 & 0.45 & 1{,}118 \tabularnewline
Premature & CMTF & 84.4 & 0.39 & 1{,}002 \tabularnewline
Risky & CMTF & 88.4 & 0.45 & 1{,}118 \tabularnewline
Cross-domain & CMTF & 88.6 & 0.46 & 1{,}119 \tabularnewline
\bottomrule
\end{tabular}
\end{table}

\begin{figure}[!t]
\centering
\includegraphics[width=0.72\linewidth]{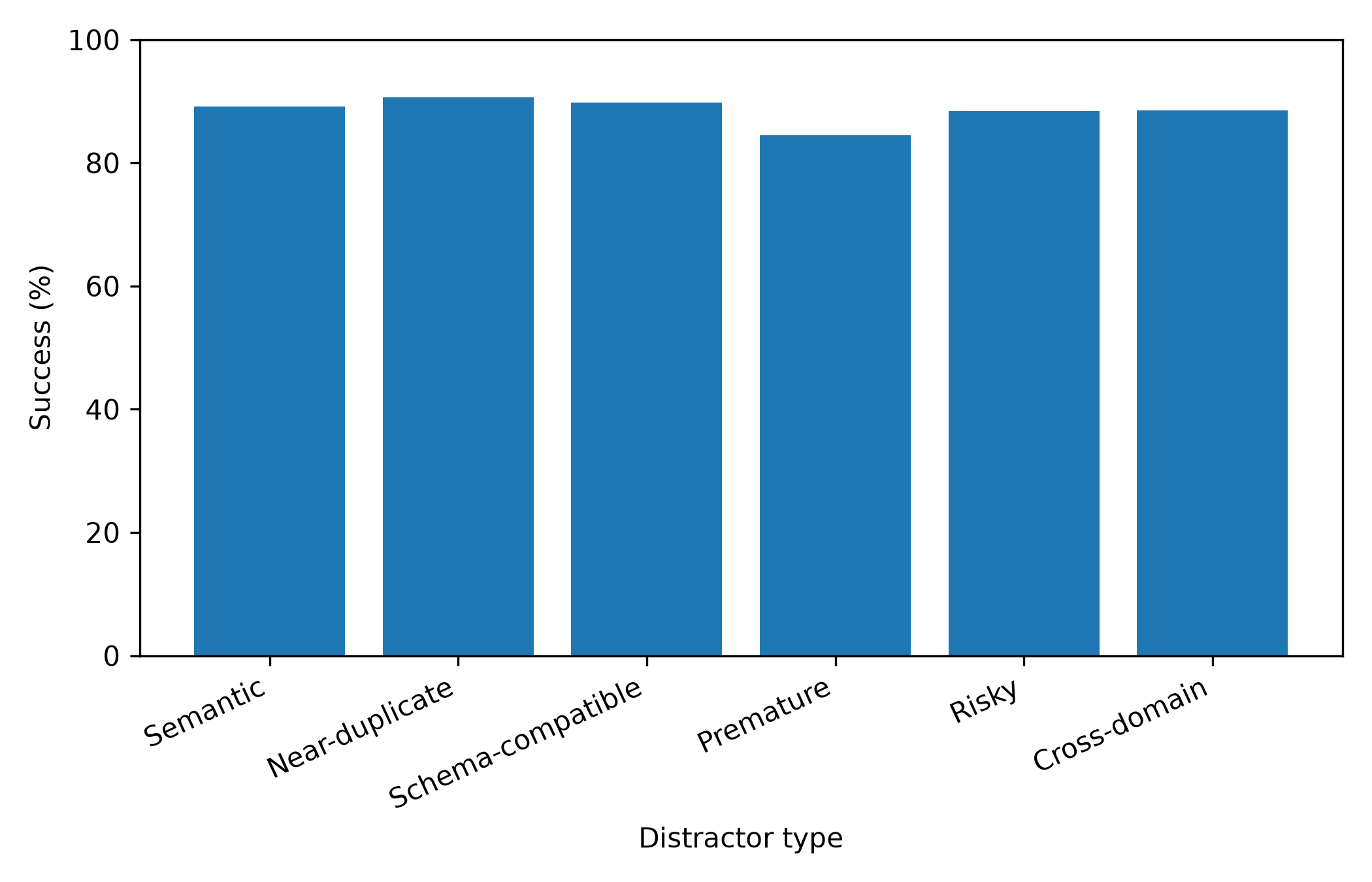}
\caption{CMTF robustness across targeted distractor types. Success remains high across semantic, near-duplicate, schema-compatible, premature, risky, and cross-domain distractors.}
\label{fig:distractor-robustness}
\end{figure}

\subsection{Causal Filtering versus Relevance and Executability Baselines}
\label{sec:results-causal}

The results highlight a distinction between relevance, executability, and causal necessity. Keyword top-$k$ methods reduce the visible menu, but they do not reason about whether a tool is valid for the current state or useful for reaching the goal. State-aware filtering requires satisfied preconditions, but executability alone is not enough: a callable tool may still be irrelevant, premature, or risky. Full causal path exposure identifies tools that may appear on a valid path, but it can still expose future actions before they are needed.

CMTF applies the strictest criterion by exposing only the next causally useful frontier. This explains why it achieves both higher success and lower token usage. The result is not simply that smaller menus are better; keyword top-5 also exposes small menus but performs poorly. The stronger conclusion is that useful tool menus must be aligned with the agent's current state and goal progress.

\subsection{Efficiency and Risk Exposure}
\label{sec:results-efficiency-risk}

Efficiency and risk exposure follow the same pattern. All-tools exposure averages 56{,}062 tokens and 206.87 risky-tool exposures per task execution. CMTF reduces these values to 1{,}125 tokens and 0.71 risky-tool exposures. This indicates that causal filtering provides a least-exposure interface: it reduces the number of irrelevant or high-impact actions shown to the model while preserving the actions needed for task completion.

This does not replace external authorization or policy enforcement. However, it reduces the burden placed on those downstream safeguards by limiting the model's opportunity to select risky or premature tools in the first place.

\subsection{Model-Level Consistency}
\label{sec:results-models}

Table~\ref{tab:model-method} reports CMTF performance by model in the main mixed-distractor benchmark. Absolute performance varies across model backends, showing that model capability still matters. However, CMTF achieves high success on several model families while keeping token usage low. The lower success rates for some models indicate that menu construction improves the agent interface but does not eliminate model-specific tool-selection errors. This variation motivates reporting both aggregate and model-level results, since menu construction and model capability interact.

\begin{table}[t]
\centering
\caption{Model-level CMTF results in the main mixed-distractor benchmark. Success is reported in percent; other metrics are averaged per task execution.}
\label{tab:model-method}
\begin{tabular}{lrrr}
\toprule
Model & Success (\%) & Wrong tools & Tokens \tabularnewline
\midrule
Claude Haiku 4.5 & 100.0 & 0.07 & 896 \tabularnewline
Claude Opus 4.8 & 100.0 & 0.00 & 1{,}107 \tabularnewline
Claude Sonnet 4.6 & 62.2 & 1.17 & 1{,}103 \tabularnewline
GPT-OSS-120B & 95.6 & 0.03 & 1{,}632 \tabularnewline
Nova 2 Lite & 100.0 & 0.00 & 858 \tabularnewline
Nova 2 Pro & 67.8 & 1.37 & 1{,}177 \tabularnewline
Nova Premier & 74.4 & 1.07 & 1{,}099 \tabularnewline
\bottomrule
\end{tabular}
\end{table}

\FloatBarrier

\section{Discussion}
\label{sec:discussion}

\subsection{Tool Menus as an Agent-Interface Problem}
\label{sec:discussion-interface}

The results support the view that tool use is not only a model capability problem, but also an agent-interface problem. A tool-augmented model does not choose from the full space of possible actions; it chooses from the menu exposed at each decision step. This menu determines the model's immediate action space, the amount of schema and description text included in the prompt, and the number of plausible but incorrect actions available to select.

This distinction is important because exposing all tools is not a neutral baseline. Although it guarantees that the correct tool is present, it also changes the agent's action space by increasing ambiguity, token cost, and risky-tool exposure. ToolMenuBench therefore treats menu construction as a first-class component of agent design. The benchmark results show that changing the visible menu can substantially change downstream task success even when the task, model, execution environment, and tool registry are held fixed.

\subsection{Filtering Is Not Just Compression}
\label{sec:discussion-compression}

A key finding is that smaller menus alone are not sufficient. Keyword top-$k$ filtering exposes far fewer tools than the all-tools baseline, but its task success remains low in the main benchmark. This shows that the benefit of CMTF is not merely prompt compression. A useful filter must preserve the next required action while removing tools that are plausible but wrong for the current state.

This separates tool filtering from retrieval-only tool selection. A tool may be textually relevant to the user request, executable under the current state, or schema-compatible with available arguments, yet still fail to advance the task. ToolMenuBench makes this distinction explicit by evaluating wrong-tool calls, premature actions, risky exposure, and success under different distractor types.

\subsection{Why Causal Filtering Helps}
\label{sec:discussion-causal}

Causal filtering helps because multi-step tool use has temporal and state-dependent structure. Many workflow tasks require actions in a specific order: searching before reading, reading before summarizing, drafting before sending, or identifying an object before updating it. Preconditions and effects provide a lightweight way to represent this structure.

Full causal path exposure uses this structure to remove irrelevant tools, but it can still expose tools that are useful later rather than useful now. CMTF applies a stricter criterion by exposing only the next causally useful frontier. This explains why it achieves both higher success and lower token usage in the main benchmark. The stronger conclusion is not that the smallest menu is always best, but that the visible menu should be aligned with the agent's current state and goal progress. This benefit depends on the availability and quality of tool contracts, so causal filtering should be viewed as an interface design principle rather than a claim that perfect contracts are always available.

\subsection{Efficiency, Risk, and Least Exposure}
\label{sec:discussion-risk}

The efficiency results show that tool filtering can reduce more than prompt length. Since tool descriptions and schemas are repeatedly included in the model context, broad tool exposure directly increases token usage. CMTF substantially reduces this cost by exposing only the tools needed for the current step.

The same least-exposure principle also matters for safety-relevant risk exposure. High-impact tools such as sending, deleting, sharing, or modifying external state should ideally be hidden or gated unless they are relevant to the current step and the required state has been reached. Causal filtering does not replace authorization checks, confirmation flows, audit logs, or policy enforcement. However, it can reduce the burden on those safeguards by limiting the model's opportunity to select risky or premature actions in the first place.

\subsection{Implications for Tool Registry Design}
\label{sec:discussion-registry}

The benchmark suggests that reliable tool registries should describe more than names, natural-language descriptions, and input schemas. They should also encode when each tool is appropriate to call, what state it requires, what state it produces, and whether it performs a high-impact action. These precondition--effect contracts make tool menus state-dependent rather than purely relevance-dependent.

This has practical implications for agent platform design. Near-duplicate tools, schema-compatible tools, and premature actions should be documented and separated carefully because they create realistic failure modes. Tool registries should include explicit contracts, risk labels, and state variables where possible. When such metadata is unavailable, learned-contract methods may help populate or audit these fields, but their reliability should be evaluated separately. The present results show that even lightweight causal structure can substantially improve the reliability and efficiency of tool-augmented agents.

\section{Limitations and Threats to Validity}
\label{sec:limitations}

ToolMenuBench uses a controlled benchmark environment to isolate the effect of tool-menu construction. This improves internal validity because each filtering method is evaluated under the same tasks, model backends, tool registries, prompting protocol, and execution rules. However, the synthetic registry and mocked execution environment do not capture all properties of production systems, including API latency, authentication failures, schema drift, permission scopes, external side effects, and ambiguous or incomplete tool documentation.

The benchmark also relies on a fixed symbolic state representation and precondition--effect contracts. This makes causal filtering directly measurable, but real agents may operate under partial observability, noisy state extraction, incomplete contracts, or non-canonical state variables. The strong performance of causal filtering should therefore be interpreted as evidence that state-aware causal structure is valuable when available, not as a claim that such structure is always easy to obtain.

The current task suite focuses on workflow domains such as calendar, email, files, documents, contacts, and identity resolution. These domains are representative of many assistant-style tool-use settings, but they do not cover all agent environments. Results may differ for open-ended web navigation, code execution, database administration, robotics, financial workflows, or long-horizon enterprise workflows with many interacting systems.

The results also depend on model-specific tool-calling behavior. ToolMenuBench evaluates the interaction between a filtering method and a model backend, not the filter in isolation. Differences in instruction following, tool-call formatting, refusal behavior, and sensitivity to prompt wording can affect downstream success even when the visible tool menu is held fixed.

The reported experiments evaluate the core benchmark setting under a controlled design, but ToolMenuBench also supports extensions that are outside the empirical scope of this paper. These include embedding-based retrieval, schema-aware filtering, learned-contract CMTF, perturbed-contract sensitivity, recovery-menu filtering, and full policy-aware risk or cost optimization.

Finally, token usage and risky-tool exposure are useful proxies for efficiency and safety-relevant exposure, but they are not substitutes for production authorization, confirmation, auditing, and policy enforcement. Future work should validate ToolMenuBench on real tool registries and staged deployment environments with imperfect contracts and richer operational constraints.

\section{Conclusion}
\label{sec:conclusion}

We introduced ToolMenuBench, a benchmark for evaluating tool-menu construction in tool-augmented LLM agents. Rather than treating the available tool set as a fixed background condition, ToolMenuBench makes the visible tool menu itself the object of evaluation. This allows agent systems to be compared not only by whether a model can call tools, but also by which tools are exposed, when they are exposed, and how exposure affects reliability, efficiency, and safety-relevant risk.

In a controlled evaluation spanning 26{,}460 end-to-end agent executions, we evaluated seven model backends, three tool-menu sizes, six filtering methods, and seven evaluation settings. The results show that tool-menu design has a substantial effect on downstream behavior. In particular, causal minimal tool filtering improves task success while reducing visible tools, token usage, wrong-tool calls, and risky-tool exposure relative to unfiltered exposure, lexical filtering, state-aware filtering, and broader causal-path baselines.

These findings suggest that scalable LLM agents require more than larger context windows or better tool-call syntax. They require principled interfaces that align the visible action space with the agent's current state, goal progress, and operational constraints. Future agent evaluations should therefore measure tool exposure as a first-class systems variable alongside task success, efficiency, and safety-relevant risk exposure.

\appendices

\section{Additional Distractor Robustness Summary}
\label{app:distractor-summary}

The main paper reports the best-performing method for each targeted distractor setting in Table~\ref{tab:distractor-results}. For completeness, Table~\ref{tab:appendix-distractor-summary} summarizes the CMTF result in each targeted distractor setting. Full per-method distractor robustness tables are included in the released artifacts.

\begin{table}[H]
\centering
\caption{CMTF results across targeted distractor settings. Success is reported in percent; wrong-tool calls and tokens are averaged per task execution.}
\label{tab:appendix-distractor-summary}
\begin{tabular}{lrrr}
\toprule
Distractor type & Success (\%) & Wrong tools & Tokens \tabularnewline
\midrule
Semantic & 89.2 & 0.44 & 1{,}108 \tabularnewline
Near-duplicate & 90.6 & 0.42 & 1{,}111 \tabularnewline
Schema-compatible & 89.8 & 0.45 & 1{,}118 \tabularnewline
Premature & 84.4 & 0.39 & 1{,}002 \tabularnewline
Risky & 88.4 & 0.45 & 1{,}118 \tabularnewline
Cross-domain & 88.6 & 0.46 & 1{,}119 \tabularnewline
\bottomrule
\end{tabular}
\end{table}

\FloatBarrier

\section*{Acknowledgment}
The authors thank colleagues for helpful feedback. This work was conducted in the authors' personal capacity. The views expressed in this paper are solely those of the authors and do not necessarily reflect the views of their employers. This work did not receive external funding. The authors declare no conflicts of interest.

\section*{Artifact Availability}

The ToolMenuBench benchmark artifacts, including the synthetic tool registry, task definitions, gold trajectories, precondition--effect contracts, distractor annotations, filtering implementations, evaluation scripts, result tables, and figure-generation utilities, are available at: \url{https://github.com/R-Suresh/ToolMenuBench}.

\bibliographystyle{IEEEtran}
\bibliography{references}

@inproceedings{yao2022react,
  title     = {{ReAct}: Synergizing Reasoning and Acting in Language Models},
  author    = {Yao, Shunyu and Zhao, Jeffrey and Yu, Dian and Du, Nan and Shafran, Izhak and Narasimhan, Karthik and Cao, Yuan},
  booktitle = {International Conference on Learning Representations (ICLR)},
  year      = {2023}
}

@inproceedings{schick2023toolformer,
  title     = {Toolformer: Language Models Can Teach Themselves to Use Tools},
  author    = {Schick, Timo and Dwivedi-Yu, Jane and Dessi, Roberto and Raileanu, Roberta and Lomeli, Maria and Hambro, Eric and Zettlemoyer, Luke and Cancedda, Nicola and Scialom, Thomas},
  booktitle = {Advances in Neural Information Processing Systems (NeurIPS)},
  year      = {2023}
}

@inproceedings{qin2023toollm,
  title     = {{ToolLLM}: Facilitating Large Language Models to Master 16000+ Real-World APIs},
  author    = {Qin, Yujia and Liang, Shihao and Ye, Yining and Zhu, Kunlun and Yan, Lan and Lu, Yaxi and Lin, Yankai and Cong, Xin and Tang, Xiangru and Qian, Bill and Zhao, Sihan and Hong, Lauren and Tian, Runchu and Xie, Ruobing and Zhou, Jie and Gerstein, Mark and Li, Dahai and Liu, Zhiyuan and Sun, Maosong},
  booktitle = {International Conference on Learning Representations (ICLR)},
  year      = {2024}
}

@inproceedings{li2023apibank,
  title     = {{API-Bank}: A Comprehensive Benchmark for Tool-Augmented {LLM}s},
  author    = {Li, Minghao and Zhao, Yingxiu and Yu, Bowen and Song, Feifan and Li, Hangyu and Yu, Haiyang and Li, Zhoujun and Huang, Fei and Li, Yongbin},
  booktitle = {Proceedings of the 2023 Conference on Empirical Methods in Natural Language Processing (EMNLP)},
  pages     = {3102--3116},
  year      = {2023}
}

@inproceedings{patil2025bfcl,
  title     = {The {Berkeley Function Calling Leaderboard} ({BFCL}): From Tool Use to Agentic Evaluation of Large Language Models},
  author    = {Patil, Shishir G. and Mao, Hanxue and Yan, Frank and Ji, Chen Cheng and Suresh, Vishnu and Stoica, Ion and Gonzalez, Joseph E.},
  booktitle = {Proceedings of the 42nd International Conference on Machine Learning},
  series    = {Proceedings of Machine Learning Research},
  volume    = {267},
  pages     = {48371--48392},
  year      = {2025}
}

@inproceedings{shi2025toolret,
  title     = {Retrieval Models Aren't Tool-Savvy: Benchmarking Tool Retrieval for Large Language Models},
  author    = {Shi, Zhengliang and Wang, Yuhan and Yan, Lingyong and Ren, Pengjie and Wang, Shuaiqiang and Yin, Dawei and Ren, Zhaochun},
  booktitle = {Findings of the Association for Computational Linguistics: ACL 2025},
  pages     = {24497--24524},
  year      = {2025}
}

@misc{liu2025toolscope,
  title         = {{ToolScope}: Enhancing {LLM} Agent Tool Use through Tool Merging and Context-Aware Filtering},
  author        = {Liu, Marianne Menglin and Garcia, Daniel and Parllaku, Fjona and Upadhyay, Vikas and Shah, Syed Fahad Allam and Roth, Dan},
  year          = {2025},
  eprint        = {2510.20036},
  archivePrefix = {arXiv},
  primaryClass  = {cs.CL},
  url           = {https://arxiv.org/abs/2510.20036}
}

@misc{babu2026toolchoice,
  title         = {{ToolChoiceConfusion}: Causal Minimal Tool Filtering for Reliable {LLM} Agents},
  author        = {Babu, Rahul Suresh and Iyer, Laxmipriya Ganesh},
  year          = {2026},
  eprint        = {2606.06284},
  archivePrefix = {arXiv},
  primaryClass  = {cs.AI},
  url           = {https://arxiv.org/abs/2606.06284}
}

@misc{babu2026contract2tool,
title         = {{Contract2Tool}: Learning Preconditions and Effects for Reliable Tool-Augmented {LLM} Agents},
author        = {Babu, Rahul Suresh and Iyer, Laxmipriya Ganesh},
year          = {2026},
eprint        = {2606.07904},
archivePrefix = {arXiv},
primaryClass  = {cs.AI},
url           = {https://arxiv.org/abs/2606.07904}
}

@misc{babu2026selfhealing,
  title         = {Self-Healing Agentic Orchestrators for Reliable Tool-Augmented Large Language Model Systems},
  author        = {Babu, Rahul Suresh and Agrawal, Adarsh},
  year          = {2026},
  eprint        = {2606.01416},
  archivePrefix = {arXiv},
  primaryClass  = {cs.AI},
  url           = {https://arxiv.org/abs/2606.01416}
}

\end{document}